\documentclass[twocolumn]{article}
\usepackage[top=20truemm,bottom=25truemm,left=20truemm,right=20truemm]{geometry}
\usepackage{graphicx,color}
\usepackage{authblk}
\usepackage{amsmath}

\title{{\bf  Control as Probabilistic Inference \\as an Emergent Communication Mechanism \\in Multi-Agent Reinforcement Learning}
}
\author[1]{Tomoaki Nakamura}
\author[2]{Akira Taniguchi}
\author[2]{Tadahiro Taniguchi}

\affil[1]{ The University of Electro-Communications}
\affil[2]{ Ritsumeikan University}

\date{}

\begin{document}
\maketitle
\begin{abstract}
This paper proposes a generative probabilistic model integrating emergent communication and multi-agent reinforcement learning. 
The agents plan their actions by probabilistic inference, called control as inference, and communicate using messages that are latent variables and estimated based on the planned actions.  
Through these messages, each agent can send information about its actions and know information about the actions of another agent. 
Therefore, the agents change their actions according to the estimated messages to achieve cooperative tasks. 
This inference of messages can be considered as communication, and this procedure can be formulated by the Metropolis-Hasting naming game. 
Through experiments in the grid world environment, we show that the proposed PGM can infer meaningful messages to achieve the cooperative task.
\end{abstract}

\thispagestyle{empty}

\section{Introduction}
Communication allows humans to engage in cooperative actions and accomplish tasks.
Furthermore, a task can be executed efficiently by creating symbols (e.g., words and gestures) that are only shared among the participants involved in the task.
In this study, we consider a symbol emergence model for a multistep cooperative task performed by two agents.

\begin{figure}[t]
	\begin{center}
	\includegraphics[scale=0.7]{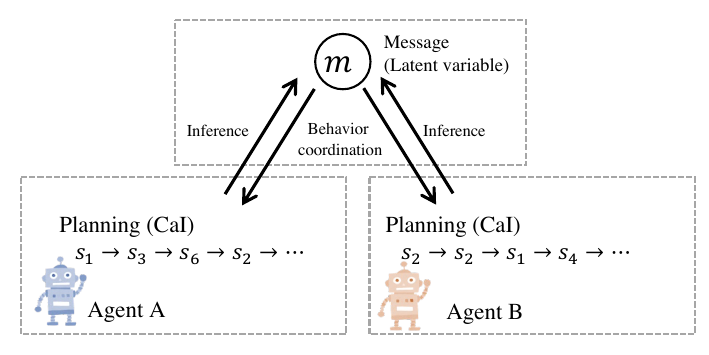}
	\caption{Overview of symbol emergence during a cooperative task between two agents.}
	\label{fig:abst}
	\end{center}
\end{figure}

In this paper, we propose a probabilistic generative model comprising a Markov decision process that determines the actions of each agent and a message that acts as a latent variable coordinating the actions of both agents.
Figure \ref{fig:abst} shows an overview of the proposed model.
Each agent plans its actions using probabilistic inference based on control as inference (CaI) framework \cite{DBLP:journals/corr/abs-1805-00909}.
A shared latent variable is inferred based on planned actions. 
Because this latent variable is shared, each agent can obtain the information of another agent and change its plan through this latent variable.
In other words, this latent variable acts as a message and the actions of both agents are coordinated by the communication that involves exchanging the messages.
To emerge the message of the cooperative task, we employed the Metropolis--Hastings naming game (MHNG) \cite{hagiwara2022multiagent,taniguchi2022emergent}, which is based on the Metropolis--Hastings algorithm.
In the original MHNG, two agents generate symbols representing the objects observed by them, that is, object category names.
In this study, we applied the MHNG to obtain symbols representing the states of the two agents, which are used for communicating their own states and understanding the states of each other. 

Deep reinforcement learning has been used for multiagent tasks \cite{sukhbaatar2016learning, foerster2016learning, jiang2018learning}.
In these studies, multiple agents were connected through a network and messages were inferred by making them differentiable variables through back propagation. In other words, the error information computed from the internal states of others is directly transmitted to oneself, which is an unnatural modeling process from a communication perspective.
By contrast, the MHNG-based method \cite{hagiwara2022multiagent,taniguchi2022emergent} avoids unnatural assumptions and composes the inferences of message variables as natural communication.
Deep learning-based models for emergent communication have also been proposed \cite{ermcom_2018_iclr, bouchacourt-baroni-2019-miss, emcom_scale}.
Most of these studies employed one-way communication, from the sender to the receiver, and included a referential game task wherein an appropriate target was selected through communication. However, these studies have not been applied to tasks that require multistep action selection through bidirectional communication, such as the task employed in this study.

\section{Proposed Method}

\begin{figure}[t]
	\begin{center}
	\includegraphics[scale=0.3]{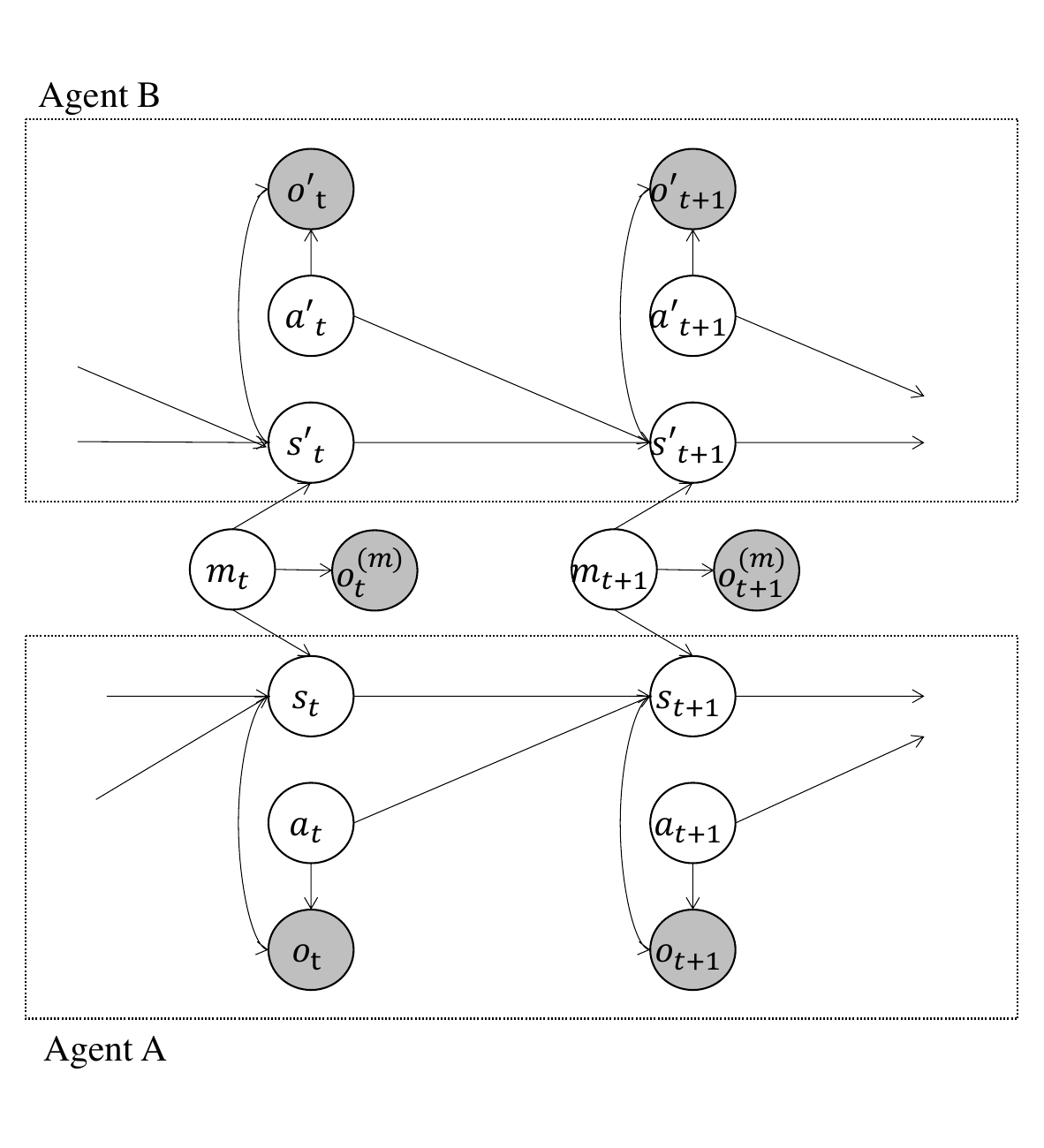}
	\caption{Graphical model of generating cooperative actions for two agents. }
	\label{fig:model}
	\end{center}
\end{figure}

Figure \ref{fig:model} shows a graphical model of the cooperative action generation of the two agents, and the details of each stochastic variable are listed in Table \ref{tbl:params}. 
It was assumed that the behavior of each agent was generated through a Markov decision process.
State $s_t$ of an agent at time $t$ is determined according to state $s_{t-1}$, action $a_{t-1}$, and message $m_t$, which is the shared latent variable:
\begin{eqnarray}
s_t \sim p(s_t | m_t, s_{t-1}, a_{t-1}). 
\end{eqnarray} 
Because this model can influence the state of others through latent variable $m_t$, this latent variable can be considered a message. Furthermore, the process of inferring the optimal value of $m_t$ from the states and actions of both agents can be considered communication, as described further on.

The optimality variable $o^{(m)}_t \in \{0, 1\}$ represents the state optimality of the two agents: 1 indicates optimal, whereas 0 indicates not optimal.
Therefore, probability $p(o^{(m)}_t=1 | m_t )$ of the optimality variable is the optimality of the states of both agents represented by message $m_t$. 

Similarly, $o_t, o'_t \in \{0, 1\}$ is the optimality variable of each agent's state and action: 1 indicates that the state and action are optimal, whereas 0 indicates they are not.
The probability $p(o_t=1 | s_t, a_t)$ of this optimality variable represents the optimality of the state and action and is assumed to be computed using reward function $r(s_t, a_t)$ as follows:
\begin{eqnarray}
p(o_t=1 | s_t, a_t ) \propto \exp( r(s_t, a_t) ). 
\end{eqnarray} 
In other words, by inferring state $s_t$ and message $m_t$ under the condition that the value of the optimality variables is always 1, the optimal state sequence for both agents can be calculated as follows: 
\begin{eqnarray}
s_t, m_t \sim p(s_t, m_t | s'_t, o_{1:T} =1, o^{(m)}_{1:T}=1  ). 
\end{eqnarray} 
\begin{table}[t]
  \caption{Parameter details. }
  \label{tbl:params}
  \centering
  \begin{tabular}{|c|l|} \hline
    $m_t$		&Message communicated between agents \\ \hline
    $o^{(m)}_t$	& Optimality variable of agent states \\ \hline
    $s_t, s'_t$	& State of each agent \\ \hline
    $a_t, a'_t$	& Action of each agent \\ \hline
    $o_t, o'_t$	& Optimality variable of \\ 
                & ~~~state and actions of each agent \\ \hline
  \end{tabular}
\end{table}
However, this equation has two problems: it includes the internal state $s'_t$ of others, which cannot be observed in practice, and it is difficult to derive this probability distribution analytically.
We solved these problems by alternately inferring the following two variables:
\begin{eqnarray}
s_t &\sim& p(s_t | o_{t:T} =1,  m_{1:T})  {\rm ~~~:planning,} \label{eq:plan} \\
m_t &\sim& p(m_t | s_t, s'_t, o^{m}_{1:T}=1) {\rm ~~:communication.} \label{eq:comm}
\end{eqnarray} 
Equation (\ref{eq:plan}) describes state planning, which can be computed based on the CaI framework \cite{DBLP:journals/corr/abs-1805-00909} proposed by Levine.
Equation (\ref{eq:comm}) describes the inference of the message and can be formulated using the MHNG proposed by Taniguchi et. al \cite{hagiwara2022multiagent,taniguchi2022emergent}, which allows both agents to infer messages through communication without observing the internal states of each other.

The optimal states and messages were inferred using the following procedure: 
\begin{enumerate}
   \item Set the distribution $p(s_t | m^*_t )$ of states given that message $m^*_t$ constitutes a uniform distribution. 
	
   \item Iterate the following steps $C$ times.
   \begin{enumerate}
	   \item Using $p(s_t | m^*_t )$, optimal states under message $m^*_t$ are inferred using the CaI framework. 
		\begin{eqnarray}
		s^*_t &\sim& p(s_t | o_{1:T} =1, m^{*}_{1:T}  ) ~t=1, \cdots, T. ~~~~
		\end{eqnarray} 
	   \item Using inferred state $s^*_t, {s'}^*_t$, the $m^*_t$ is updated through the MHNG. 
		\begin{eqnarray}
		m^*_t \sim p(m_t | s^*_t, {s'}^*_t, o^{m}_{1:T}=1) ~~~t=1, \cdots, T. 
		\end{eqnarray} 
   \end{enumerate}
\end{enumerate}
where $C$ denotes the number of times both the agents communicate.
In subsequent sections, the equations required for each inference are derived.

\subsection{Task Planning Using Messages}
\subsubsection{Computation of Backward Probability}

The probability that the value of optimality variables $o_{t:T}$ is 1 at all future times under message $m_{t:T}$ can be expressed as 
\begin{eqnarray}
&& p(o_{t:T}=1 | s_t, a_t, m^*_{t:T}) \nonumber \\
&&~~~~~ = p(o_t=1 | s_t, a_t)  p( o_{t+1:T}=1| s_t, a_t, m^*_{t:T})  \\
&&~~~~~= p( o_t=1| s_t, a_t ) \int p( o_{t+1:T}=1 | s_{t+1}, m^*_{t+1:T} )  \nonumber \\
&&~~~~~~~~~~  p(s_{t+1} | s_t, a_t, m^*_{t+1} ) ds_{t+1}   \\
&&~~~~~ \equiv q(s_t, a_t),  \label{eq:backward}
\end{eqnarray}
where $p(s_{t+1}| s_t, a_t, m^*_{t+1} ) \propto p(s_{t+1}|s_t, a_t) p(s_{t+1}|m^*_{t+1})$ is obtained using the product of the expert approximation, and the equation is transformed to 
\begin{eqnarray}
q(s_t, a_t) &\approx& p( o_t=1| s_t, a_t ) \int p( o_{t+1:T}=1 | s_{t+1}, m^*_{t+1:T} ) \nonumber \\
&&~~~~~~  p(s_{t+1}|s_t, a_t) p(s_{t+1}|m^*_{t+1}) ds_{t+1}. \label{eq:q} 
\end{eqnarray}
Next, we define $v(s_t)$ as 
\begin{eqnarray}
v(s_t) &=& p( o_{t:T}=1 | s_{t}, m^*_{t:T} ) \\
  &=& \int p(o_{t:T}=1 | s_t, a_t, m^*_{t:T}) p(a_t | s_t) da_t \\
  &=& \int q(s_t, a_t) p(a_t | s_t) da_t. \label{eq:v} 
\end{eqnarray}
By using $v(s_t)$, Eq. (\ref{eq:q}) becomes 
\begin{eqnarray}
q(s_t, a_t) &=& p( o_t=1| s_t, a_t ) \int v(s_{t+1}) \nonumber \\
&&   p(s_{t+1}|s_t, a_t) p(s_{t+1}|m^*_{t+1})ds_{t+1}. 
\end{eqnarray}
Using these results, $v(s_T)$ can be computed using $q(s_T, a_T)$, and $q(s_{T-1}, a_{T-1})$ can be computed using $v(s_T)$.
Therefore, by starting the computation from $q(a_T, s_T)$, we can compute $q(a_t, s_t)$ and $v(s_t)$ at all times as follows: 
\begin{eqnarray}
&&q(a_T, s_T) \rightarrow v(s_T) \rightarrow q(a_{T-1}, s_{T-1}) \nonumber \\
&&~~~~~~~ \rightarrow v(s_{T-1}) \rightarrow q(a_{T-2}, s_{T-2}) \cdots  \rightarrow q(a_1, s_1) \nonumber \label{eq:recur}
\end{eqnarray}

\subsubsection{Computation of Forward Probability}
The probability of a state under all past optimality variables is 1 and can be calculated as follows:
\begin{eqnarray}
\alpha(s_t) &=& p(s_t | o_{1:t-1}=1, m^*_{1:t-1}) \\
&\approx&  \int \int p(s_t | s_{t-1}, a_{t-1} ) \nonumber \\ 
&&~~~  p(a_{t-1}|s_{t-1}, o_{t-1}=1) p(s_{t-1} | m^*_{t-1} )  \nonumber \\
&&~~~  p(s_{t-1} |o_{1:t-2}=1, m^*_{1:t-2} ) ds_{t-1} da_{t-1} \\
&=&  \int \int p(s_t | s_{t-1}, a_{t-1} ) p(a_{t-1}|s_{t-1}, o_{t-1}=1)   \nonumber \\
&&~~~~~  p(s_{t-1} | m^*_{t-1} )  \alpha(s_{t-1}) ds_{t-1} da_{t-1}.  \label{eq:forward}
\end{eqnarray}
Assuming that $p( a_{t-1}| s_{t-1} )$ is uniformly distributed, 
\begin{eqnarray}
p(a_{t-1}|s_{t-1}, o_{t-1}=1) \propto p(o_{t-1}=1| a_{t-1}, s_{t-1} ), 
\end{eqnarray}
and Eq. (\ref{eq:forward}) becomes 
\begin{eqnarray}
\alpha(s_t) \propto  \int \int p(s_t | s_{t-1}, a_{t-1} )  p(o_{t-1}=1| a_{t-1}, s_{t-1} )  \\
 p(s_{t-1} | m^*_{t-1} )  \alpha(s_{t-1}) ds_{t-1} da_{t-1}. 
\end{eqnarray}
Therefore, by starting the computation from $\alpha(s_1) = p(s_1)$, we can compute $\alpha(s_t)$ at all times, from $t=1$ to $t=T$. 

\subsubsection{Optimal State Distribution}
Using the backward and forward probabilities computed above, the probability of the state wherein the value of all optimality variables is 1 at all times is computed as follows: 
\begin{eqnarray}
&& p(s_t | o_{1:T} =1,  m^*_{1:T}) \\
&&~~~~ \propto  p( o_{t:T}=1 | s_{t}, m^*_{t:T} )  \nonumber \\
&&~~~~~~~~ p(s_t | o_{1:t-1}=1, m^*_{1:t-1})  p(o_{1:t-1}=1 ) \\
&&~~~~ \propto v(s_t) \alpha(s_t). 
\end{eqnarray} 
This equation can be solved without using the other agent's state $s'_t$; therefore, each agent can compute the optimal state distribution using the received message and its own internal parameters.

\subsection{Message Generation for Cooperation}

The inferred state $s^*_t, {s'}^*_t$ is then used to infer message $m^*_t$ whose optimality variable $o^{(m)}_t$ = 1:
\begin{eqnarray}
m^*_t \sim p(m_t | s^*_t, {s'}^*_t, o^{(m)}_t=1). \label{eq:msg}
\end{eqnarray} 
However, this equation includes the internal state $s'_t$ of the other states, which cannot be directly observed.
Therefore, we considered inference using the Metropolis-Hastings algorithm, similar to \cite{hagiwara2022multiagent,taniguchi2022emergent}.
First, to generate samples that follow Equation (\ref{eq:msg}), the target distribution is defined as 
\begin{eqnarray}
&& p(\hat{m}|s^*_t, {s'}^*_t, o^{(m)}_t=1)  \nonumber \\
&&~~~~~~~ = \frac{ p(s^*_t | \hat{m} ) p({s'}^*_t, o^{(m)}_t=1 | \hat{m} ) p(\hat{m}) }{ p(s^*_t) p({s'}^*_t) } \\
&&~~~~~~~ \propto p( \hat{m}| s^*_t ) p( \hat{m} | {s'}^*_t, o^{(m)}_t=1), \label{eq:target_dist}
\end{eqnarray} 
where prior distributions $p(s^*_t), p({s'}^*_t), p( \hat{m}| s^*_t ) $ are set as uniform distributions.
When Agent A decides to accept or reject a message from Agent B, the proposed distribution for Agent B is obtained as follows: 
\begin{eqnarray}
Q(\hat{m} | m ) = p( \hat{m} | {s'}^*_t, o^{(m)}_t=1)  \label{eq:prop_dis}
\end{eqnarray} 
Using the target distribution from Equation (\ref{eq:target_dist}) and the proposed distribution from Equation (\ref{eq:prop_dis}), the acceptance probability $r$ of message $\hat{m}$ can be computed as follows: 
\begin{eqnarray}
r &=& \frac{ p( \hat{m} |s^*_t, {s'}^*_t, o^{(m)}_t=1)  Q(m|\hat{m})  }{ p(m |s^*_t, {s'}^*_t, o^{(m)}_t=1)  Q( \hat{m}| m) }  \\
&=&  \frac{ p(\hat{m}|s^*_t) p(\hat{m}|{s'}^*_t, o^{(m)}_t=1)  p( m | {s'}^*_t, o^{(m)}_t=1) }{ p(m|s^*_t) p(m|{s'}^*_t, o^{(m)}_t=1)  p( \hat{m} | {s'}^*_t, o^{(m)}_t=1)   }  \nonumber \\ \\
&=&  \frac{ p(\hat{m}|s^*_t)  }{ p(m|s^*_t)  }. 
\end{eqnarray} 
In other words, the probability $r$ that Agent A accepts the proposed message can be calculated using only the parameters of Agent A.
By iterating the process of proposing and accepting/rejecting messages, both agents can infer message $m^*$ that follows the target distribution (Eq. (\ref{eq:target_dist})) according to the Metropolis-Hastings algorithm.

\section{Experiments}
To verify that the proposed method could generate cooperative behavior, a two-agent movement task in a $2 \times 4$ grid world was conducted.

\subsection{Experimental Setup}
\begin{figure}[t]
	\begin{center}
	\includegraphics[scale=0.3]{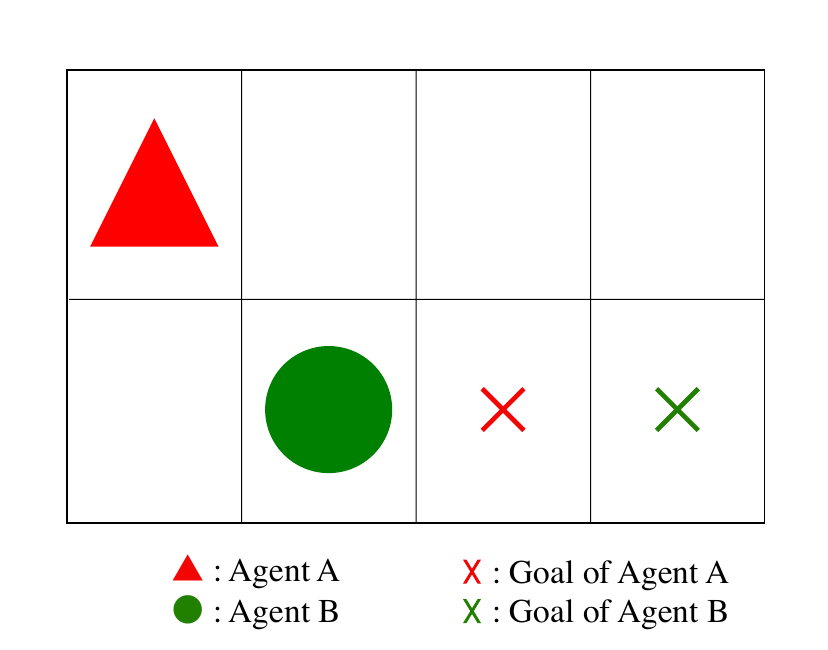}
	\caption{Experimental environment.}
	\label{fig:world}
	\end{center}
\end{figure}
\begin{figure}[t]
	\begin{center}
	\includegraphics[scale=0.6]{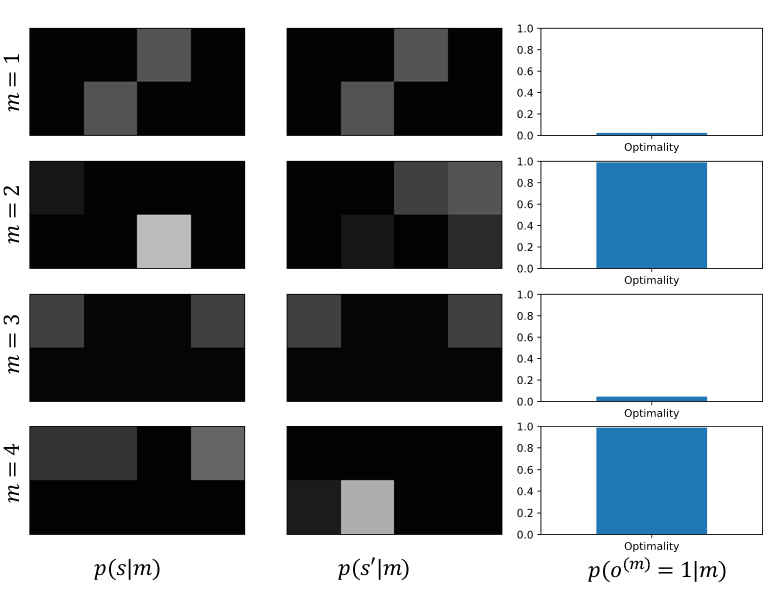}
	\caption{Learned messages. }
	\label{fig:msg}
	\end{center}
\end{figure}
\begin{figure*}[t]
	\begin{center}
	\includegraphics[scale=0.8]{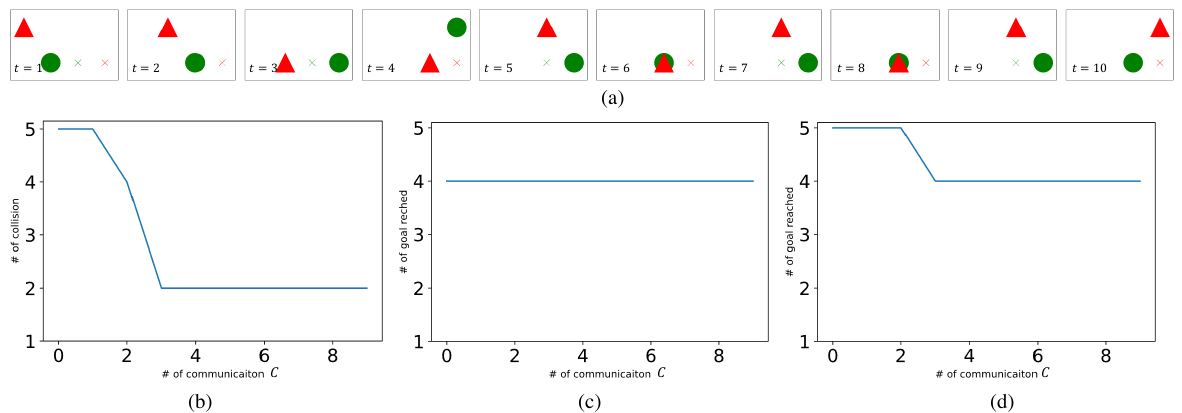}
	\caption{(a) Actions inferred after communicating and planning $C=10$ times, (b) number of agent collisions, (c) number of times Agent A achieved the goal, (d) number of times Agent B achieved the goal.}
	\label{fig:collision_num}
	\end{center}
\end{figure*}
Figure {\ref{fig:world}} shows the $2 \times 4$ grid world used in the experiment.
The triangles and circles represent agents, whose aim is to achieve their respective goals without colliding with each other.
To plan the goal position, the probability distribution representing individual optimality was set as follows:
\begin{eqnarray}
&&p(o_t=1 | s_t, a_t ) \nonumber   \\
&&~~~ \propto   \begin{cases}
        1   &   \text{: if the agent can reach the goal}  \\
            &   \text{~~~by taking action $a_t$ at $s_t$}     \\
        10^{-7}        &   \text{: otherwise} 
    \end{cases}  \nonumber \\
\end{eqnarray} 
The optimality variable $o^{(m)}_t$ of cooperative behavior was set to zero if the agents collided, and one otherwise.
The agents could perform four actions:  moving up, down, left, and right, but could not remain in the same grid.
In other words, to obtain larger accumulated optimality variables, the agents were required to continue moving while avoiding collisions and repeatedly enter and leave the goal grid as many times as possible.

States $s_t, s'_t$ denote the grid indices, $m_t$ denotes 32-dimensional categorical variables, and a multinomial distribution is used as their distribution.
The distribution of optimality $o_t, o'_t, o^{(m)}_t \in \{0, 1\}$ was binomial.

\subsection{Message Learning}

The probability distribution parameters in the message model were learned from the optimality variables of cooperative behavior, actions, and states of both agents moving randomly over 200 steps in the grid world. 
Message $m_t$ was inferred using the MHNG. A part of probability distributions ($p(s|m), p(s'|m), p(o^{(m)}|m)$) represented by the learned messages is shown in Fig. \ref{fig:msg}.
It was correctly learned that the optimality was low for $m=1$ and $m=3$, which indicated that both agents were in the same location, and high for $m=2$ and $m=4$, which indicated that the agents were in different locations.
These results show that messages expressing both states were learned through communication.

\subsection{Action Planning}
Thereafter, we tested whether the learned parameters can be used to achieve the goal without a collision.
The goal position is denoted by x in Fig.\ref{fig:world}, where the lower left grid is the origin $(0, 0)$, and $(0, 2)$ and $(0, 3)$ are the goals of Agents A and B, respectively. 
Communication and planning were iterated $C=10$ times and a $T=10$ step path was planned.
The planned actions are shown in Figure \ref{fig:collision_num}(a); although there were two collisions, the actions required to avoid the collisions and to repeat entering and leaving for each goal were planned.
With respect to the number of communication repetitions, times the agents collided, and times they reached the goal in the planned 10-step path are shown in Figures \ref{fig:collision_num}(b)--(d), respectively. 
In the case of $C=0$, wherein the agents did not communicate, the number of times each agent reached the goal was high, but collisions occurred at 5 out of 10 steps.
In the case of $C=3$, wherein the agents communicated three times, the number of times Agent B reached the goal decreased to four, but the number of collisions decreased to two. 
This indicates that the agents could plan a path that allowed them to reach their goals while avoiding collisions.

\section{Conclusions}
We proposed a probabilistic generative model that can learn and generate cooperative behavior through symbol emergence by integrating CaI and MHNG.
Through a simple experiment in a grid world, we demonstrated that the proposed model can generate messages for cooperation and realize cooperative behavior.

This paper is preliminary, and we are planning the following future work: 
\begin{itemize}
      \item Use continuous variables as states and actions combining deep reinforcement learning.
      \item Use continuous variables as messages using Gaussian process latent variable models or variational auto-encoder
      \item Formulate the PGM for cooperative tasks of more than three agent
      \item Extend the proposed method to PoMDP by introducing the state space model
\end{itemize}

\section*{Acknowledgment}
This work was supported by the JST Moonshot R\&D program, Grant Number JPMJMS2011.

\footnotesize
\bibliographystyle{unsrt}
\bibliography{reference}

\end{document}